\documentclass[10pt,twocolumn,letterpaper]{article}

\usepackage{iccv}
\usepackage{times}
\usepackage{epsfig}
\usepackage{graphicx}
\usepackage{amsmath}
\usepackage{amssymb}
\usepackage{multirow}
\usepackage{booktabs}
\usepackage{soul}
\usepackage{subfigure}

\usepackage[pagebackref=true,breaklinks=true,letterpaper=true,colorlinks,bookmarks=false]{hyperref}

\iccvfinalcopy 

\def\iccvPaperID{1756} 

\begin{document}

\title{Vision-Language Transformer and Query Generation for Referring Segmentation}
\author{
Henghui Ding\footnotemark[1]
\quad
Chang Liu\footnotemark[1]
\quad
Suchen Wang
\quad
Xudong Jiang\\
Nanyang Technological University, Singapore\\
{\tt\small \{ding0093, liuc0058, wang.sc, exdjiang\}@ntu.edu.sg}
}
\maketitle
\renewcommand{\thefootnote}{\fnsymbol{footnote}}
\footnotetext[1]{Equal contribution}


\begin{abstract}
   In this work, we address the challenging task of referring segmentation. The query expression in referring segmentation typically indicates the target object by describing its relationship with others. Therefore, to find the target one among all instances in the image, the model must have a holistic understanding of the whole image. To achieve this, we reformulate referring segmentation as a direct attention problem: finding the region in the image where the query language expression is most attended to. We introduce transformer and multi-head attention to build a network with an encoder-decoder attention mechanism architecture that ``queries'' the given image with the language expression. Furthermore, we propose a Query Generation Module, which produces multiple sets of queries with different attention weights that represent the diversified comprehensions of the language expression from different aspects. At the same time, to find the best way from these diversified comprehensions based on visual clues, we further propose a Query Balance Module to adaptively select the output features of these queries for a better mask generation. Without bells and whistles, our approach is light-weight and achieves new state-of-the-art performance consistently on three referring segmentation datasets, RefCOCO, RefCOCO+, and G-Ref. Our code is available at~\href{https://github.com/henghuiding/Vision-Language-Transformer}{https://github.com/henghuiding/Vision-Language-Transformer}.
\end{abstract}

\section{Introduction}

\begin{figure}[t]
   \begin{center}
      \includegraphics[width=1\linewidth]{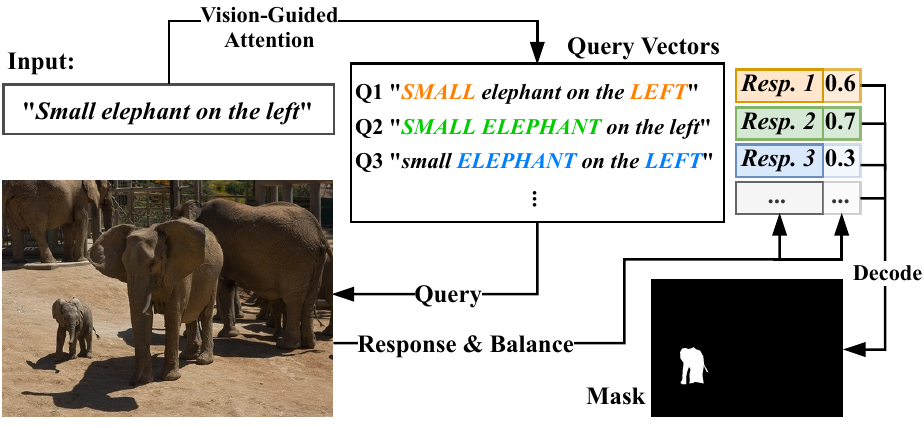}
   \end{center}
    \vspace{-0.1in}
   \caption{Our method detects multiple emphasis or understanding ways for one language expression, and produces a query vector for each of them. We use each vector to ``query'' the image, generating a response to each query. Then the network selectively aggregates these responses, in which queries that provide better comprehensions are spotlighted.}
   \label{fig:fig1}
\end{figure}

Referring segmentation targets to generate segmentation mask for the target object referred by a given query expression in natural language \cite{hu2016segmentation,liu2017recurrent,li2018referring,ding2020phraseclick}. As the referring segmentation involves both natural language processing and computer vision, it is considered as one of the most fundamental and challenging multi-modal tasks. With the recent success of learning methods, a lot of deep-learning-based works are proposed in this area and have achieved remarkable performance. However, there are still many challenges left in this task. The objects in images of referring segmentation are correlated in a complicated manner while the query expression frequently indicates the target object by describing the relationships with others, which requires a holistic understanding on the image and language expression. Another challenge is caused by the varieties of objects/images as well as the unrestricted expression of languages, which brings a high degree of randomness. 


Firstly, to address the challenge of complicated correlations in the given image and language, 
we explore to enhance the holistic understanding of multi-modal features by building the network with global operations, in which direct interactions are modeled among all elements (\eg, pixel-pixel, word-word, pixel-word). Currently, the Fully Convolutional Network (FCN)-like framework~\cite{long2015fully, ding2018context, ding2019semantic, ding2019boundary} is widely used in referring segmentation methods~\cite{hu2016segmentation,margffoy2018dynamic}. They usually perform convolution operations on the fused (\eg, concatenated) vision-language features to generate the target mask. However, the long-range dependencies modeling in regular convolution operation is indirect, as its large receptive field is typically achieved by stacking small-kernel convolutions. This oblique process brings inefficiencies to information interactions among pixels/words in a distance~\cite{wang2018non}, thus is undesirable for referring segmentation models to understand the global context of the image~\cite{ye2019cross}. In recent years, the attention mechanism is gaining respectable popularity in the computer vision community for its advantage in building direct interaction among all elements, which greatly helps the model in capturing the global semantic information. Some previous referring segmentation works also use attention to alleviate the long-range dependency issues~\cite{ye2019cross,hu2020bi,shi2018key}. However, most of them only utilize the attention mechanism as auxiliary modules based on the FCN-like pipeline, which limits their ability to model the global context. %
In this work, we reformulate the referring segmentation problem in terms of attention and reconstruct the current FCN-like framework with Transformer~\cite{vaswani2017attention}. We generate a set of query vectors from language features using vision-guided attention, and use these vectors to ``query'' the given image and generate the segmentation mask from the responses, as shown in Fig.~\ref{fig:fig1}.~This attention-based framework enables us to implement global operation among multi-modal features in every stage of computation, making the network better at modeling the global context of both vision and language information.

Secondly, to deal with the randomness caused by the varieties of objects/images and the unconstrained expression of languages, we propose to comprehend the language expression in different ways incorporating with vision features. In many previous referring segmentation methods, such as~\cite{luo2020multi, yu2018mattnet}, the language self-attention is often used to extract the most informative part and emphasized word(s) in the language expression. For these methods, their linguistic understanding is derived alone from the language expression itself without interacting with the image, so that they cannot distinguish which emphasis is more suitable and effective that better fit a particular image. Thus their detected emphases might be inaccurate or inefficient.
On the other hand, in most previous vision-transformer works, queries for the transformer decoders are usually a set of fixed learned vectors, each of which 
is used to predict one object. Experiments show that each query vector 
has its own operating modes and specializes in certain kinds of objects~\cite{carion2020end}. In these works with fixed queries, there must imply a hypothesis that objects in the input image are distributed under some statistic rules, which does not match the randomness of referring segmentation. 
To address these issues, we propose a Query Generation Module (QGM) to produce multiple different query vectors based on the language, and with the aid of vision features. Each vector comprehends the language expression in its own way. 
With the proposed QGM, we improve the diversity of ways to understanding the image and query language, enhancing the network's robustness in dealing with highly random inputs. At the same time, to ensure the generated query vectors are valid and find the more suitable comprehension ways to the image and language, we further propose a Query Balance Module to adaptively select the output features of these queries for a better mask generation. 

Our approach builds deep interactions between language and vision features at different levels, greatly enhancing the fusion and utilization of multi-modal features. Besides, the proposed module is lightweight and its parameter size is roughly equivalent to seven convolution layers. In summary, our main contributions are listed as follows:
\begin{itemize}
\setlength\itemsep{0em}
   \item We
   design a Vision-Language Transformer~(VLT) method to build deep interactions among multi-modal information and enhance the holistic understanding to vision-language features.
   \item We propose a Query Generation Module that understands the language from different comprehension ways, and a Query Balance Module to focus on the suitable ways.
   \item The proposed method achieves new state-of-the-art on multiple datasets consistently, especially on hard and complex datasets.
\end{itemize}

\section{Related Works}

\begin{figure*}[t]
   \begin{center}
      \includegraphics[width=0.97\linewidth]{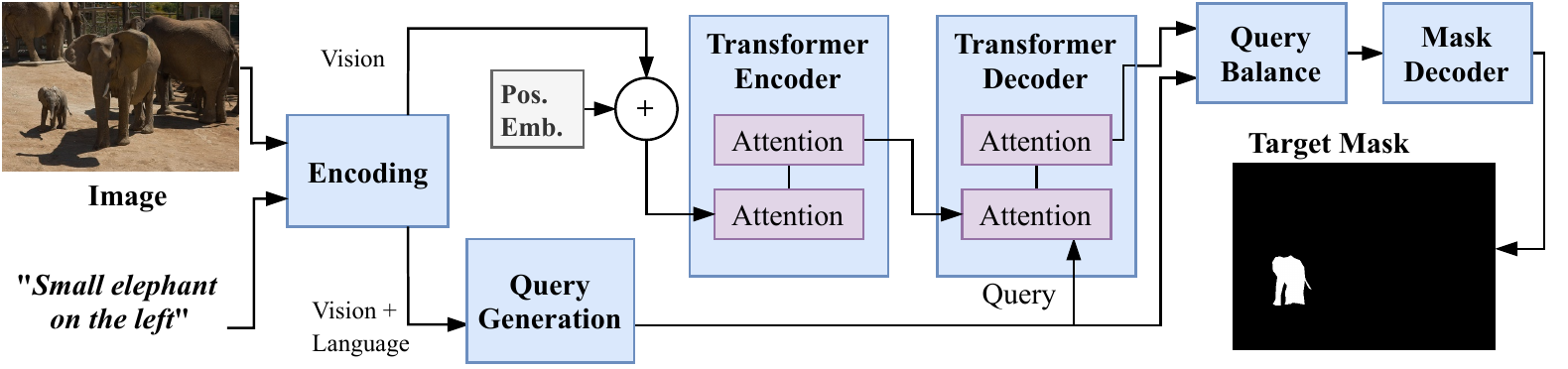}
   \end{center}
   \caption{The overview of the network framework. Firstly, the input image and language expression are transformed into feature spaces. Features then processed by a transformer encoder-decoder model, generating a set of query responses. These responses are then decoded to output the target mask. ``Pos. Emb.'': Positional Embeddings.}
   \label{fig:network}
\end{figure*}
\subsection{Referring Segmentation}

The aim of the referring segmentation is to find the target object in an image given an natural language expression describing its properties. The task is first proposed by Hu \etal in~\cite{hu2016segmentation}, in which a set of fused features are extracted by concatenating the linguistic features extracted by LSTM and vision features extracted by CNN. Then, a fully convolutional network (FCN) is applied on the fused feature, proving the feasibility of this problem. In~\cite{liu2017recurrent}, in order to make use of each word in the referring sentence, Liu \etal proposed a recurrent network based on multimodal LSTM (mLSTM). The framework model each word in every recurrent stage, so that the language feature is better fused with vision features along all the forward process. In~\cite{yu2018mattnet}, Yu \etal proposed a two-stage method which first extract multiple instances using an instance segmentation netwrok Mask R-CNN~\cite{he2017mask} then utilizes language features to select the target from the extracted instances. Luo \etal~\cite{luo2020multi} proposed a framework which jointly learns to solve two tasks: referring expression comprehension (REC)~\cite{luo2017comprehension, hu2017modeling} and segmentation (RES), achieving remarkable performance. Besides, with the attention-based method getting into people's sight, researchers find that the attention mechanism preciously suit the formulation of the task. This is shown by a number of works, such as~\cite{ye2019cross} designed a Cross-Modal Self-Attention (CMSA) model which adaptively focus on informative words in the query expression and the important part of the input image, and~\cite{hu2020bi} utilizes a pair of attention module namely language-guided visual attention module and vision-guided linguistic attention module to learn the relationship between multi-modal features.

Unlike previous methods that are built upon FCN-like networks, we replace the prediction and identification head with a fully attention based architecture, which helps us to easily model the long-range dependencies in the image.

\subsection{Attention and Transformer}

The Transformer model, which is a sequence to sequence deep network architecture that uses only the attention mechanisms, is first introduced by Vaswani \etal in~\cite{vaswani2017attention}. The transformer model quickly gain its attraction in the Natural Language Processing (NLP) and shows promising performance on several major NLP tasks such as machine translation~\cite{vaswani2017attention}, language modeling~\cite{krause2019dynamic}, question answering~\cite{devlin2018bert}. In recent years, the transformer is also being adopted in the computer vision community and has shown potential on various tasks such as object detection~\cite{carion2020end}, image recognition~\cite{dosovitskiy2020image}, human-object interaction~\cite{wangsuchen_iccv2021}, semantic segmentation~\cite{zheng2020rethinking}, etc. Unlike CNN that focus on local pixels (kernels), transformer is appreciated because its ability on modeling global information.

\section{Methodology}

The overall architecture of our approach is shown in Fig.~\ref{fig:network}.  The network takes an image $I\in R^{H'\times W' \times 3}$ and an language expression $T=\left\{w_i\right\}_{i=1,\dots ,t}$ as input, where $H'$ and $W'$ are the height and width of the input image respectively, $t$ is the length of the language expression. Firstly, the input image and language expression are mapped into the feature space. Next, language and vision features are processed together by the Query Generation Module (QGM) to produce a set of language query vectors, which represent different comprehensions about the image and language expression. At the same time, vision features are sent to the transformer encoder to generate a set of memory features. The query vectors obtained from QGM are used to ``query'' the memory features, and the resulting responses from decoder are then selected by a Query Balance Module. Finally, the network outputs a mask $M_p$ for the target object.

\begin{figure}[t]
   \begin{center}
      \includegraphics[width=0.9\linewidth]{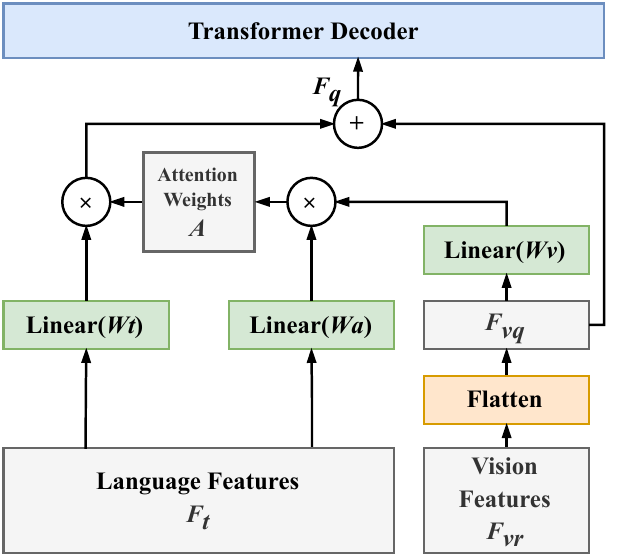}
   \end{center}
    \vspace{-0.1in}
   \caption{Architecture of the Query Generation Module. The module takes language features $F_t$ and vision feature $F_{vr}$ as input, and generate a set of query vectors $F_q$.}
   \vspace{-0.1in}
   \label{fig:query_gen}
\end{figure}
\subsection{Query Generation Module}

In most previous vision-transformer works~\cite{carion2020end}, queries for the transformer decoder are usually a set of fixed learned vectors, each of which is used to predict one object and has its own operating mode, \eg, specializes in objects of certain kinds or located in certain regions. In these works with fixed queries, there must imply a hypothesis that objects in the input image are distributed under some statistic rules. This is proved to work in other related tasks such as object detection and panoptic segmentation.

For referring segmentation, the target-of-interest indicated by the input language can be any instance in the image. Since both image and language expression are unconstrained, the randomness of the target object's properties is significantly high. Thus, fixed query vectors, like in most other vision-transformer works, are not enough to represent the properties of the target object. Instead, these properties are hidden in the language expression, \eg, keywords like ``big/small'', ``left/right''. To extract the key information and address this high randomness in referring segmentation, we propose a Query Generation Module to adaptively produce the query vectors online according to the input image and language expression with the help of image information, as shown in Fig.~\ref{fig:query_gen}. Also, in order to let the network learn different aspects of information and enhance the robustness of the queries, we generate multiple queries though there is only one target instance.

The Query Generation Module takes language feature $F_{t}\in R^{N_l\times C}$ and raw vision feature $F_{vr}\in R^{H\times W\times C}$ as input. In $F_{t}$, the $i$-th vector is the feature vector of the word $w_i$, which is the $i$-th word in the input language expression. $N_l$ in $F_{t}$ is fixed by zero-padding. The module aims to output $N_q$ query vectors, each of which is a language feature with different attention weights guided by vision information.

\begin{figure}[t]
   \begin{center}
      \includegraphics[width=1\linewidth]{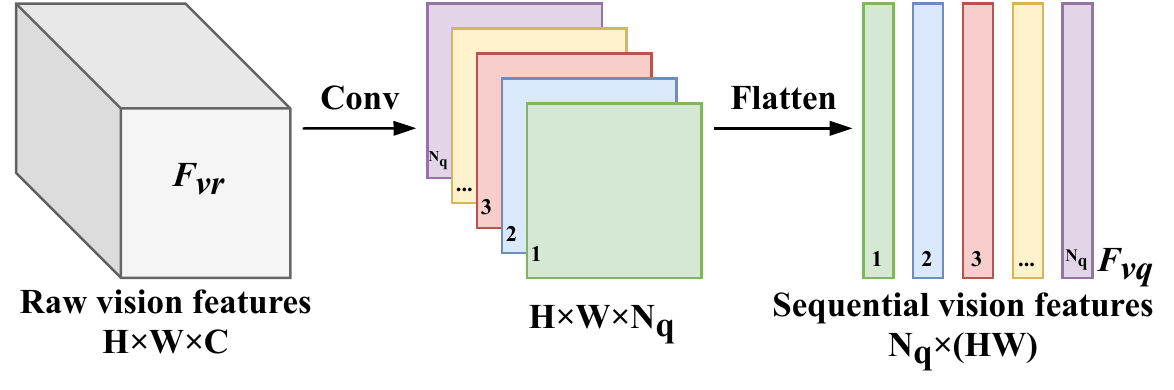}
   \end{center}
   \vspace{-0.1in}
   \caption{The preparation process of vision features in the Query Generation Module. The module converts regular 2D vision feature into a set of sequential features.}
   \vspace{-0.1in}
   \label{fig:fv_prep}
\end{figure}

Firstly, the vision features are prepared as shown in Fig.~\ref{fig:fv_prep}. We reduce the feature channel dimension size of the vision feature $F_{vr}$ to query number $N_q$ by three convolution layers, resulting in $N_q$ feature maps. Each of them will participate in the generation of one query vector. The feature maps are then be flattened in the spatial domain, forming a feature matrix $F_{vq}$ of size $N_q\times (HW)$, \ie,
\begin{equation}
   F_{vq}=\text{Flatten}(\text{Conv}(F_{vr}))^T
\end{equation}

\begin{figure}[t]
   \begin{center}
      \includegraphics[width=0.8\linewidth]{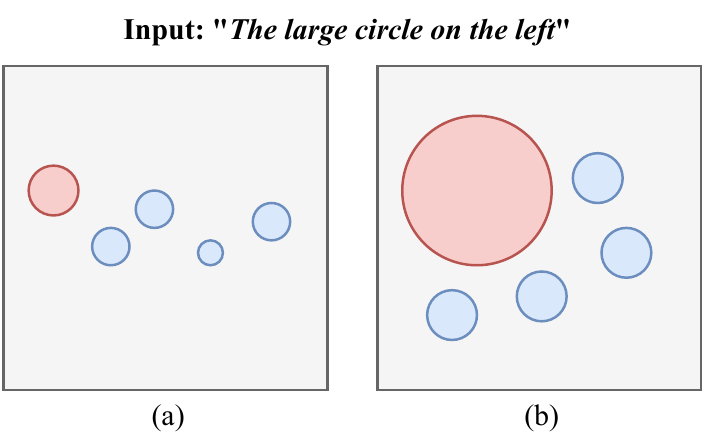}
   \end{center}
   \vspace{-0.1in}
   \caption{An example of one sentence having different emphasis. For different images, the informative degree of words ``large'' and ``left'' are different.}
   \vspace{-0.1in}
   \label{fig:q_eg}
\end{figure}

It is known that for a language expression, the importance of different words is different. Some previous works address this issue by measuring the importance of each word. For example,~\cite{luo2020multi} gives each word a weight and~\cite{yu2018mattnet,huang2020referring} defines a set of labels, \eg, location, attribute, entity, and finds the degree of each word belongs to different labels. Most works derive the weights by the language self-attention, which does not utilize the information in the image and only outputs one set of weights. But in practice, a same sentence may have different understanding perspectives and emphasis, and the most suitable and effective emphasis can only be known with the help of the image. We give an intuitive example in Fig.~\ref{fig:q_eg}. For the same input sentence ``The large circle on the left'', the word ``left'' is more informative for the first image but the ``large'' is more useful for the second image. In this case, language self-attention cannot differentiate the importance between ``large'' and ``left'' and hence can only give both words high attention weights, making the attention process less effective. Therefore, in the Query Generation Module, we comprehend the language expression from multiple aspects incorporating the image, forming $N_q$ queries from language. Different queries emphasize different words, and more suitable attention weights are then be found and enhanced by a Query Balance Module.

\begin{figure*}[t]
   \begin{center}
      \includegraphics[width=0.92\linewidth]{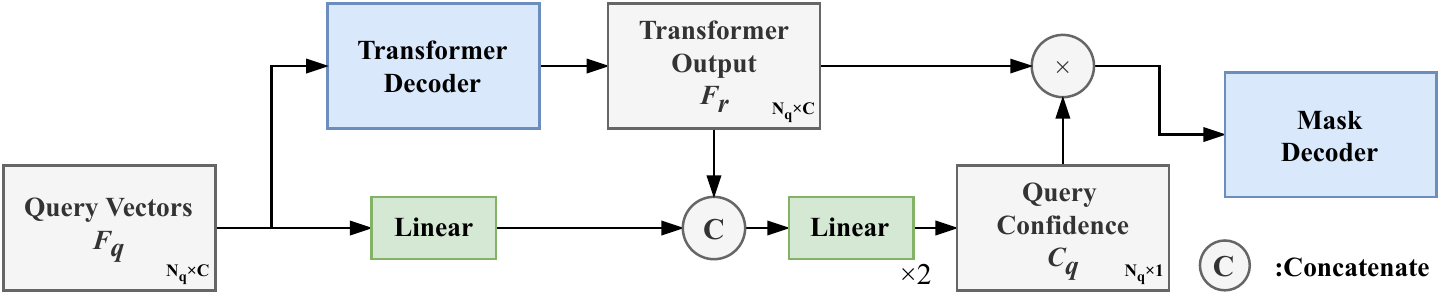}
   \end{center}
    \vspace{-0.1in}
   \caption{The Query Balance Module. A confidence parameter is computed for each query vector. The confidences are then applied on its corresponding transformer outputs to control the influence of each query vector.}
   \label{fig:query_weight}
   \vspace{-0.1in}
\end{figure*}

To this end, we derive the attention weights for language features $F_t$ by incorporating the vision features $F_{vq}$. Firstly, we apply linear project on $F_{vq}$ and $F_{t}$. Then, for the $n$-th query, we take the $n$-th vision feature vector $f_{vqn}\in R^{1\times (HW)}, n=1,2,\dots,N_q$ and language feature of all words. Let the feature of the $i$-th word denote as $f_{ti}\in R^{1\times C}, i=1,2,\dots,N_l$. The $n$-th attention weight for the $i$-th word is the product of projected $f_{vqn}$ and $f_{ti}$:
\begin{equation}
   a_{ni}=
   {\sigma(f_{vqn}W_v)\ \sigma(f_{ti}W_a)^T}
\end{equation}
resulting in a scalar $a_{ni}$. $W_v\in R^{(HW)\times C}$ and $W_a\in R^{C\times C}$ are learnable parameters and $\sigma$ is activation function. Softmax function is applied across all words for each query as normalization.

For the $n$-th query, a set of attention weights for all words are formed up from $a_{ni}$ to $A_{n}\in R^{1\times N_l}, n=1,2,\dots,N_q$. It consists of a set of attention weights for different words, and the different queries may attend to different parts of the language expression. Thus, $N_q$ query vectors focus on different emphasis, or different comprehension ways, of the language expression. 

Next, the derived attention weights are applied on the language features:
\begin{equation}
   F_{qn}=A_n \sigma(F_{t}W_t)
\label{Eq:Fq}
\end{equation}
where $W_t\in R^{C\times C}$ are learnable parameters. Each $F_{qn}$ is an attended language feature vector guided by vision information and serves as one query vector to the transformer decoder. Mathematically, each query is a projected weighted sum of features of different words in the language expression, therefore it remains properties as a language feature, and can be used to query the image.

\begin{table*}[t]
   \centering
   \small
   \caption{Comparison with Convolutional Networks in terms of parameter size and performance. The ``\#params'' represents the number of trainable parameters in our Transformer and its substitute, a module with seven 3$\times$3 convolutional layers.}
   \setlength{\tabcolsep}{3.2mm}{\begin{tabular}{l|c|c|c|c|c|c|c}
      \hline
      Type          & \#params     & IoU   &Pr@0.5 & Pr@0.6 & Pr@0.7 & Pr@0.8 & Pr@0.9 \\
      \hline \hline
      7 Conv Layers &$\sim 16.6$M & 44.28 & 49.54 & 42.16  & 35.24  & 25.98  & 10.47 \\
      Transformer   &$\sim 17.5$M & 49.36 & 55.84  & 50.79  & 41.68  & 29.96  & 10.76 \\
      \hline
   \end{tabular}}%
   \vspace{-0.05in}
   \label{tab:param}%
\end{table*}%

\begin{table*}[t]
   \centering
   \small
   \caption{Comparison of our query generation method with other related methods. ``$F_t$'': use the language features of all words as queries. ``Learnt'': queries are parameters learnt in training while fixed in testing, similar with~\cite{carion2020end}.}
   \setlength{\tabcolsep}{4.6mm}{\begin{tabular}{r|l|c|c|c|c|c|c}
      \hline
      No. & Method  & IoU   & Pr@0.5 & Pr@0.6 & Pr@0.7 & Pr@0.8 & Pr@0.9 \\
      \hline \hline
      1   & $F_t$   & 45.05 & 52.69  & 46.08  & 36.20  & 20.97  & 3.42   \\
      2   & Learnt  & 42.99 & 49.85  & 42.38  & 31.52  & 17.14  & 2.41   \\
      \hline \hline
      3   & Ours    & 49.36 & 55.84  & 50.79  & 41.68  & 29.96  & 10.76   \\
      \hline
   \end{tabular}}
   \vspace{-0.1in}
   \label{tab:exp_qgm}%
\end{table*}%

\subsection{Query Balance Module}
We get $N_q$ different query vectors from the proposed Query Generation Module. Each query represents a specific comprehension of the input language expression. As we discussed before, both the input image and language expression are of high randomness. Thus, it is desired to adaptively select the better comprehension ways and let the network focus on the more reasonable and suitable comprehension ways.
On the other hand, as the independence of each query vector is kept in the transformer decoder~\cite{carion2020end} but we only need one mask output, it is desired to balance the influence of different queries to the final output. Therefore, we propose a Query Balance Module to adaptively assign each query vector a confidence measure that reflects how much it fits the prediction and the context of the image. The architecture is shown in Fig.~\ref{fig:query_weight}.

The Query Balance Module takes the query vectors $F_q$ from the Query Generation Module and its response from the transformer decoder, $F_r$, which is of the same size as $F_q$. Let $F_{rn}$ represent the corresponding response to $F_{qn}$. In the Query Balance Module, the query and its corresponding response are first concatenated together. Then, a set of query confidence levels $C_q$ of size $N_q\times 1$ is generated by two consecutive linear layers. Each scalar $C_{qn}$ shows how much the query $F_{qn}$ fits the context of its prediction, and controls the influence of its response $F_{rn}$ to the mask decoding. The second linear layer uses sigmoid as an activation function to control the output range. Each response $F_{rn}$ is multiplied with the corresponding query confidence $C_{qn}$, and sent for mask decoding.

\subsection{Network Architecture}

\textbf{Encoding.}~Since the transformer architecture only accepts sequential inputs, the original image, and language input must be transformed into feature space before sending to the transformer. For vision features, following~\cite{carion2020end}, we use a CNN backbone for image encoding. We take the features of the last three layers in the backbone as the input for our encoder. By resizing the three sets of feature maps to the same size and summing them together, we get the raw vision feature $F_{vr}\in R^{H\times W \times C}$, where $H, W$ is the spatial size of features, and $C$ is the feature channel number. For language features, we first use a lookup table to convert each word into word embeddings~\cite{zhanghui2021}, and then utilize an RNN module to convert the word embedding to the same channel number as the vision feature, resulting in a set of language features $F_t\in R^{N_l\times C}$. $F_{vr}$ and $F_t$ are then sent to the Query Generation Module as vision and language features. At the same time, we flatten the spatial domain of $F_{vr}$ into a sequence, forming the vision feature $F_v\in R^{N_v \times C}, N_v=H\times W$, which will be sent to the Transformer Module.

\textbf{Transformer Module.}~We use a complete but shallow transformer to apply the attention operations on input features. The network has a transformer encoder and a decoder, each of which has two layers. Each layer has one (encoder) or two (decoder) multi-head attention modules and one feed-forward network, as defined in~\cite{vaswani2017attention}. The transformer encoder takes the vision feature $F_{v}$ as input, deriving the memory features about vision information $F_m\in R^{N_v\times C}$. Before sending to the encoder, we add a fixed sine spatial positional embedding on $F_{v}$. In our experiments, we then multiply $F_{v}$ with the final state of the language features as in \cite{luo2020multi} to enrich the information in vision features.  $F_m$ is then sent to the transformer decoder as keys and values, together with $N_q$ query vectors produced by the Query Generation Module. The decoder queries the vision memory feature with language query vectors and outputs $N_q$ responses for mask decoding.

\textbf{Mask Decoder Module.}~The Mask Decoder consists of three stacked $3\times 3$ convolution layers for decoding followed by one $1\times 1$ convolutional layer for outputting the final segmentation mask. Upsampling layers can be optionally plugged in between layers to control the output size. To demonstrate the effectiveness of the transformer module more clearly, in our implementation, the Mask Decoding Module does not use any former CNN features. We use the Binary Cross Entropy loss on the output mask to guide the network training.

\section{Experiments}

\subsection{Implementation Details}

\textbf{Experiment Settings.}~We strictly follow previous works~\cite{luo2020multi,yu2018mattnet} for experiment settings, including preparing the Darknet-56 backbone as the CNN encoder. Input images are resized to $416\times 416$. Each Transformer block has 8 heads, and the hidden layer size in all heads is set to 256. The maximum length for the input language expression is set to 15 for RefCOCO and RefCOCO+ and 20 for G-Ref. We train the network for 50 epochs using the Adam optimizer with the learning rate $\lambda=0.001$. With the shallow transformer architecture, we are able to train the model with a large batch size of 32 per GPU with 32GB VRAM.

\textbf{Metrics.} We use two metrics for experiments: mask IoU and Precision@$X$. The IoU metrics show the quality of the output mask which reflects the overall performance of the methods, including both targeting and mask generating abilities. The Precision@$X$ reports the successful targeting rate at the IoU threshold $X$, which focuses on the targeting ability of the method.

\begin{table*}[t!]
   \centering
   \small
   \caption{Influence of Query Numbers. $^*$: without Query Balance Module}
   \setlength{\tabcolsep}{4.6mm}{\begin{tabular}{r|c|c|c|c|c|c}
      \hline
      $N_q$ & IoU   & Pr@0.5 & Pr@0.6 & Pr@0.7 & Pr@0.8 & Pr@0.9 \\
      \hline \hline
      1     & 44.83 & 50.17  & 43.94  & 34.75  & 21.64  & 4.66   \\
      2     & 47.07 & 52.85  & 47.31  & 39.66  & 28.90  & 8.30   \\
      4     & 46.79 & 53.06  & 47.54  & 40.38  & 28.23  & 8.92   \\
      8     & 49.04 & 55.57  & 50.58  & 44.24  & 32.99  & 12.62  \\
      16    & \textbf{49.36} & \textbf{55.84}  & 50.79  & 41.68  & 29.96  & 10.76   \\
      32    & 49.27 & 55.57  & 50.48  & 44.43  & 33.87  & 12.50  \\
      \hline \hline
      16$^*$ & 48.94 & 55.41  & 50.32  & 43.84  & 32.56  & 12.99  \\
      \hline
   \end{tabular}}
   \label{tab:exp_nq}%
\end{table*}%

\begin{table*}[t!]
   \centering
   \small
   \caption{Experimental results of the IoU metric, and comparison of other methods with ours. U: UMD split. G: Google split.}
   \setlength{\tabcolsep}{2.8mm}{\begin{tabular}{l|c|c|c|c|c|c|c|c|c}
      \hline
      \multirow{2}[4]{*}{} & \multicolumn{3}{c|}{RefCOCO} & \multicolumn{3}{c|}{RefCOCO+} & \multicolumn{3}{c}{G-Ref} \\
      \cline{2-10}
                                     & val   & test A & test B & val   & test A & test B & val (U)   & test (U)  & val(G)   \\
      \hline
      DMN~\cite{margffoy2018dynamic} & 49.78 & 54.83 & 45.13 & 38.88 & 44.22 & 32.29 & -     & -     & 36.76 \\
      RRN~\cite{li2018referring}     & 55.33 & 57.26 & 53.93 & 39.75 & 42.15 & 36.11 & -     & -     & 36.45 \\
      MAttNet~\cite{yu2018mattnet}   & 56.51 & 62.37 & 51.70 & 46.67 & 52.39 & 40.08 & 47.64 & 48.61 & -     \\
      CMSA~\cite{ye2019cross}        & 58.32 & 60.61 & 55.09 & 43.76 & 47.60 & 37.89 & -     & -     & 39.98 \\
      BRINet~\cite{hu2020bi}         & 60.98 & 62.99 & 59.21 & 48.17 & 52.32 & 42.11 & -     & -     & 48.04 \\
      CMPC~\cite{huang2020referring} & 61.36 & 64.53 & 59.64 & 49.56 & 53.44 & 43.23 & -     & -     & 39.98 \\
      LSCM~\cite{hui2020linguistic}  & 61.47 & 64.99 & 59.55 & 49.34 & 53.12 & 43.50 & -     & -     & 48.05 \\
      MCN~\cite{luo2020multi}        & 62.44 & 64.20 & 59.71 & 50.62 & 54.99 & 44.69 & 49.22 & 49.40 & -     \\
      CGAN~\cite{luo2020cascade}     & 64.86 & 68.04 & 62.07 & 51.03 & 55.51 & 44.06 & 51.01 & 51.69 & 46.54 \\
      \hline
      VLT~(ours) & \textbf{65.65} & \textbf{68.29} & \textbf{62.73} & \textbf{55.50} & \textbf{59.20} & \textbf{49.36} & \textbf{52.99} & \textbf{56.65} & \textbf{49.76} \\
      \hline
      Prec@0.5 & 76.20 & 80.31 & 71.44 & 64.19 & 68.40 & 55.84 & 61.03 & 60.24 & 56.65 \\
      \hline
   \end{tabular}}%
  \vspace{-0.1in}
   \label{tab:results}%
\end{table*}%

\subsection{Datasets}

We evaluate our approach on three commonly used datasets: RefCOCO, RefCOCO+~\cite{yu2016modeling} and G-Ref~\cite{mao2016generation, nagaraja2016modeling}.

\textbf{RefCOCO \& RefCOCO+}~(UNC/UNC+)~\cite{yu2016modeling} are two of the largest and most commonly used datasets for referring segmentation. The RefCOCO dataset has 19,994 images with 142,209 referring expressions for 50,000 objects while the UNC+ dataset contains 19,992 images with 141,564 expressions for 49,856 objects.~Some kinds of words, \eg, words about absolute locations, are ``forbidden'' in the RefCOCO+ dataset, thus it is considered to be more challenging than the RefCOCO dataset.

\textbf{G-Ref}~\cite{mao2016generation, nagaraja2016modeling} is another commonly used dataset. It contains 26,711 images with 104,560 expressions referring to 54,822 objects. Compared to RefCOCO and RefCOCO+, the G-Ref has a longer average sentence length and richer word usage. Notably, there are two partitionings for this dataset: the Google partitioning~\cite{mao2016generation} and the UMD partitioning~\cite{nagaraja2016modeling}. The UMD partitioning has both validation set and test set, but the test set of the Google partitioning is not publicly released. We report the performance of our method on both kinds of partitioning.

\subsection{Ablation Study}

\begin{figure}[t!]
   \begin{center}
      \includegraphics[width=0.75\linewidth]{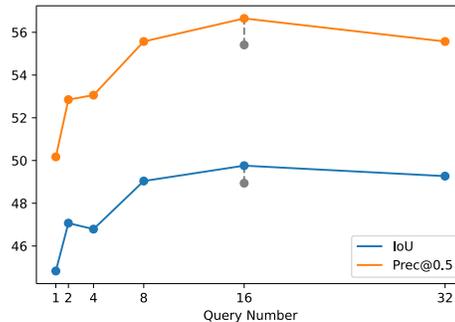}
   \end{center}
   \vspace{-0.2in}
   \caption{Performance gain by increasing the query number $N_q$. The gray points are performance without the Query Balance Module (QBM).}
   \vspace{-0.11in}
   \label{fig:nq}
\end{figure}

To better demonstrate the performance of our model on hard and complex scenarios, we do the ablation study on a more difficult dataset, the testB split of the RefCOCO+.

\textbf{Parameter Size.} We show that only a tiny transformer network can be an alternative to convolution networks while achieving better performance in our framework. In order to show the scale of our network and prove the effectiveness of the transformer module, we compare the performance and parameter size of our method with regular conv-nets in Table~\ref{tab:param}.
In the experiment, we replace the whole attention-based modules, including the transformer module, the Query Generation Module, and the Query Balance Module with stacked $3\times 3$ convolution layers that have similar parameter sizes. It shows that the parameter size of our attention-based module only roughly equivalent to 7 convolutional layers while having a much superior performance. The transformer module outperforms the 7 Conv module with over 5\% margin in IoU, and 7\% margin in Prec@0.5. This proves the effectiveness of the transformer module.


\begin{figure*}[t!]
   \begin{center}
      \includegraphics[width=0.96\linewidth]{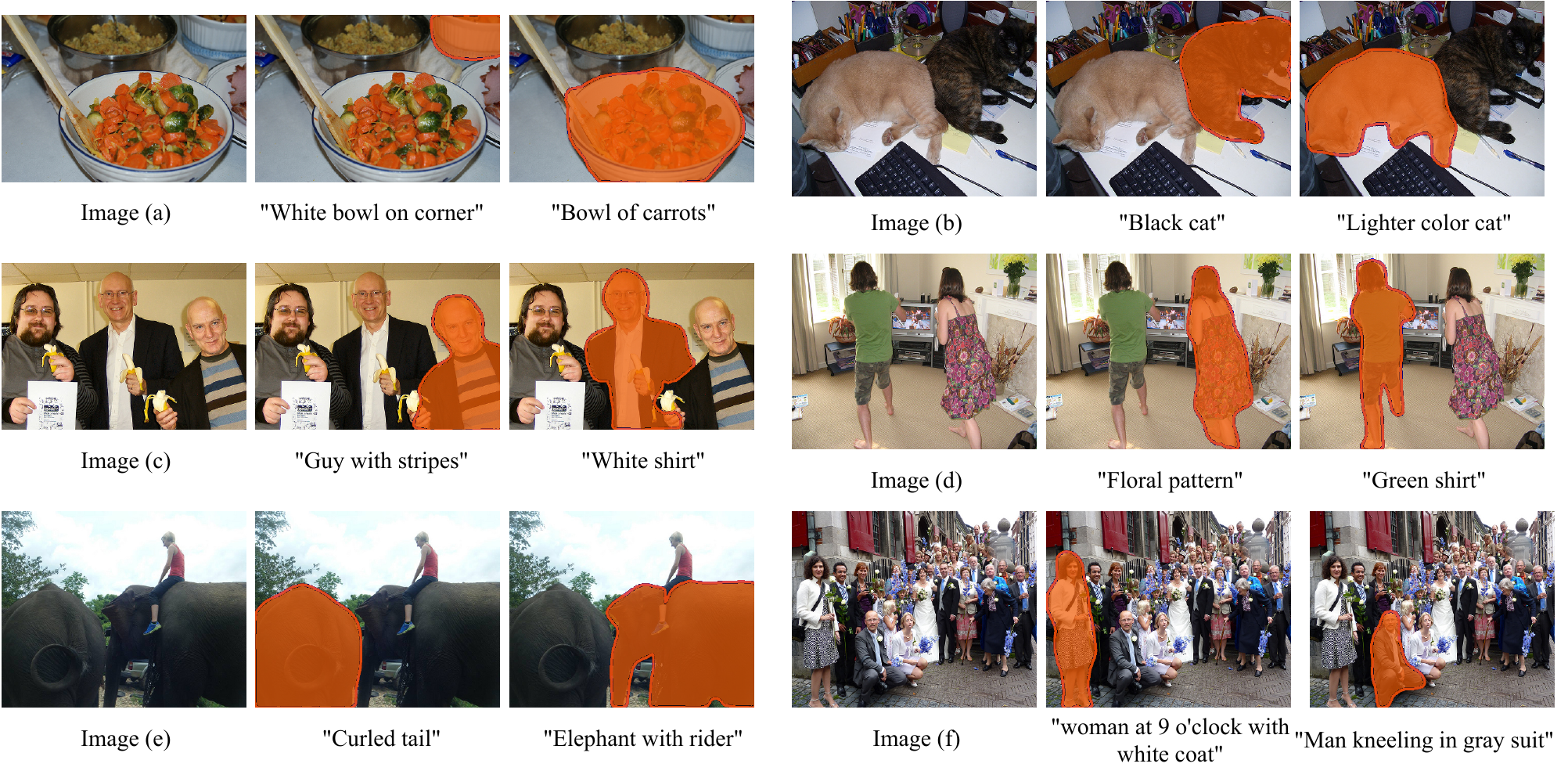}
   \end{center}
   \vspace{-0.2in}
   \caption{(Best viewed in color) Example outputs. For each set of images, the first one shows the input image, and captions under other images shows the input language expressions.}
  \vspace{-0.1in}
   \label{fig:demo}
\end{figure*}

\textbf{Query Generation.} In this section, we compare the Query Generation Module with other methods for generating query vectors. The results are reported in Table~\ref{tab:exp_qgm}. The Query Generation Module outperforms both methods with a large margin at about $3\%$ - $6\%$. In the first experiment, we directly send the language features into the transformer decoder as the query. Specifically speaking, the input language expression is processed by an RNN network, then the output for each word is used as one query vector. It can be seen that because that information between words is not sufficiently exchanged, its performance is not so satisfying. This shows that the Query Generation Module effectively understands the sentence and generates valid attended language features guided by vision information. Secondly, we use the most common method, \ie the query vectors are learned during training and fixed during inference. In the experiment, at the beginning of the training, we set 16 query vectors that are initialized with uniform distribution, and train them together with other parts of the network. It is shown that the learned fixed query vector cannot represent the target object as effectively as online produced queries by the Query Generation Module. 

\textbf{Query Number.} To show the influence of the query number $N_q$, we report the network's performance with different numbers of queries in Fig.~\ref{fig:nq} and Table.~\ref{tab:exp_nq}. The result shows that though only one mask is output, multiple queries are still desired for the transformer network. From the results, a larger query number brings a notable performance gain of about 5\% from 1 query to 16 queries. Though the IoU performance of 4 queries is slightly lower than 2 queries, from the Pr@0.5 it can be seen that its detection performance is still higher. The performance gain slows down after the query number is larger than 8, thus we choose 16 as the default query number in our implementation. This also shows that multiple queries generated by the Query Generation Module represent different aspects of information. Also, when the Query Balance Module is removed, there is a performance loss of $\sim1\%$, proving the effectiveness of the Query Balance Module.

\subsection{Comparison with State-of-the-art}

In Table~\ref{tab:results}, we compare our proposed Vision-Language Transformer (VLT) approach with previous state-of-the-art methods on three widely-used datasets.
It can be seen that our method outperforms other methods on all datasets. On the RefCOCO dataset, the IoU performance of our method is higher than other methods with $\sim$1\% gain. Furthermore, on the more difficult and complex dataset RefCOCO+, our method achieves a more notable performance gain of around 5\%, especially on the testB split. On another hard dataset G-Ref where the average length of language expression is longer, our method also achieves a higher IoU with a margin of about 2\%-5\%. This shows that the proposed approach has good abilities on hard cases and long expressions. We assume the reason is that, on the one hand, long and complex sentences usually contain more information and more emphasis, and our Query Generation and Balance Module can detect multiple emphasis and find the more informative ones. On the other hand, harder cases also may contain complex scenarios where needs a global view, and the multi-head attention is more suitable for this problem as a global operator.

\subsection{Visualization and Qualitative Results}

\begin{figure}[t]
   \centering
   \subfigure[]{
      \includegraphics[height=0.7in]{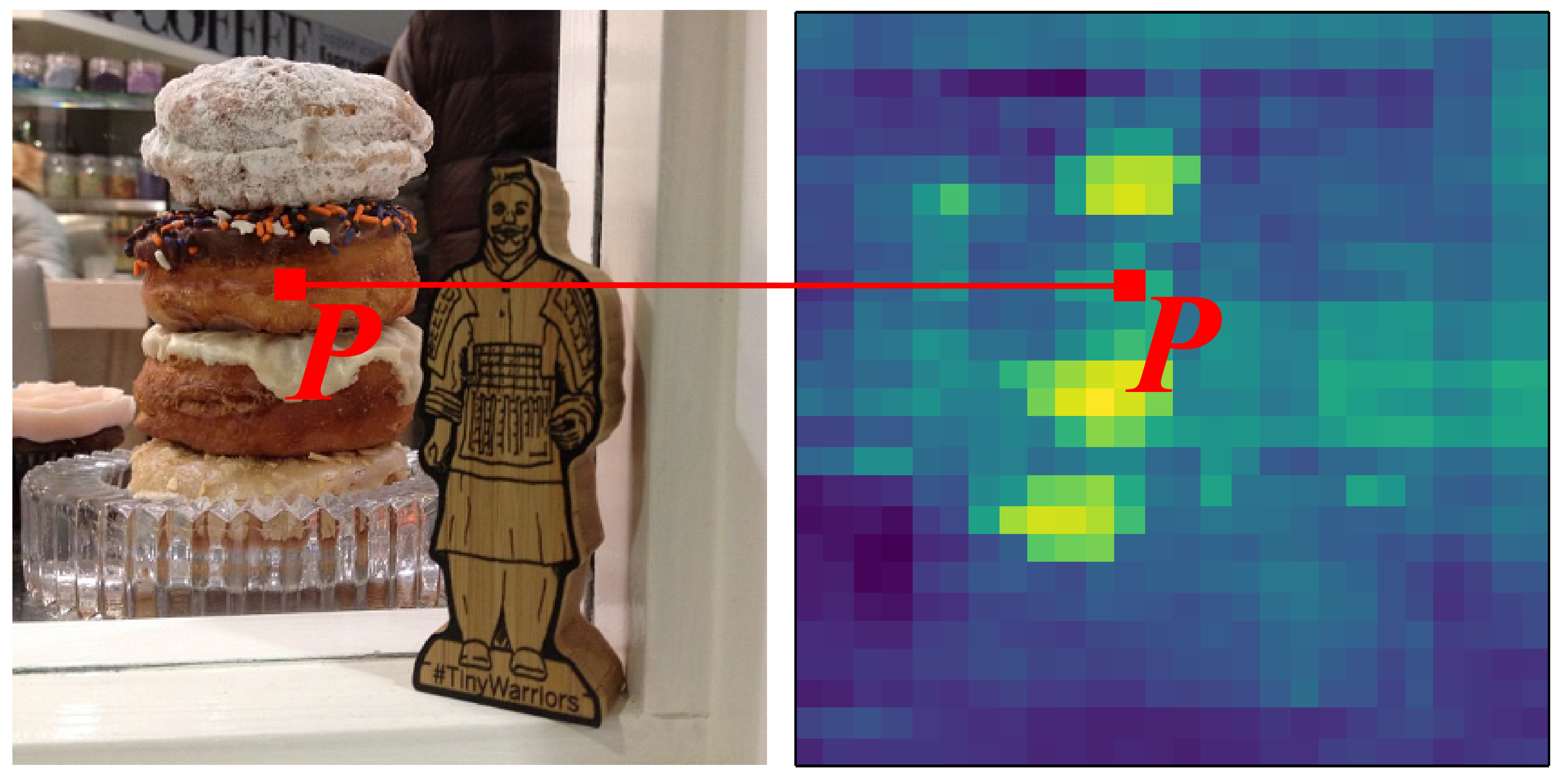}
      \label{fig:visA}
   }
   \subfigure[]{
      \includegraphics[height=0.72in]{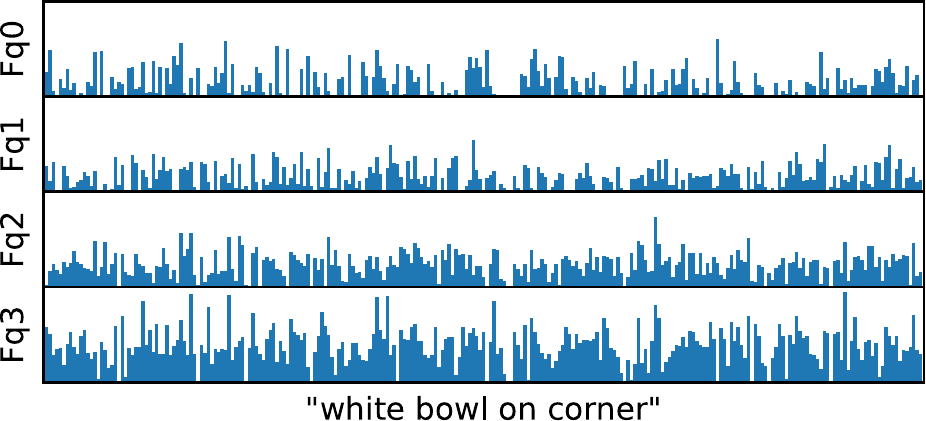}
      \label{fig:visQ}
   }
   \caption{Visualizations of: (a). the attention map of point $P$ in the transformer encoder; (b). different query vectors $F_q$.}
   \vspace{-10pt}
\end{figure}

We demonstrate example outputs of the method in Fig.~\ref{fig:demo}. To clearly show the identifying ability of the method, for each example set, we show the segmentation results of one image with different input query expressions. Image (a) and (b) are two direct cases where the language expression describes the location or color of the target. From the second expression of Image (b), it can be seen that our method is able to process the comparative words (lighter). Image (c) and (d) show the method's ability on understanding the attribute words such as ``stripes'' and relatively rarer words, \eg ``floral''. In Image (e), the method successfully identifies the target by expression describing the interaction between objects, \ie ``Elephant with rider''. The Image (f) is a photo of a group of people, where all instances distribute densely in the image in a complicated layout. Our method still managed to extract the target with difficult language expressions that contain multiple aspects of information, such as direction (9 o'clock), attributes (white coat) and posture (kneeling).


Next, we extract an attention map from the second layer of the transformer encoder of one point, as shown in Fig.~\ref{fig:visA}. It can be seen that the point from one instance attends to other related instances across the image, showing that the transformer successfully extracts long-range dependencies in one single layer. Fig.~\ref{fig:visQ} shows some query vectors $F_q$ (see Fig.~\ref{fig:query_gen} and Eq.~(\ref{Eq:Fq})), which illustrates the diversity of query vectors.

\section{Conclusion}

In this paper, we address the difficult task of referring segmentation by using the attention networks to alleviate the global information exchange problem in conventional convolutional networks. We reformulate the task to an attention problem and propose a framework that exploits the transformer to perform attention operations. To solve the problem of ambiguous referring sentence due to the unknown emphasis, we propose a Query Generation Module and a Query Balance Module to comprehend the referring sentence with the help of the referred image. The proposed model outperforms other methods with a large margin on three widely used datasets.

{\small
    \bibliographystyle{ieee_fullname}
    \bibliography{egbib}
}

\end{document}